\documentclass[11pt,letterpaper]{article}
\usepackage{ijcnlp2017}
\usepackage{times}
\usepackage{latexsym}
\usepackage{tikz}
\usepackage{CJK}
\usepackage{amsmath}
\usepackage{amssymb}
\DeclareMathOperator*{\argmax}{argmax}
\usetikzlibrary{fit,positioning,arrows.meta}
\usepackage{float}
\usepackage{tabularx}

\usepackage{multirow}

\usepackage{hyperref}

\usepackage[inline, shortlabels]{enumitem}

\tikzset{neuron/.style={shape=circle, minimum size=1.25cm, 
  inner sep=0, draw, font=\small}, io/.style={neuron, fill=gray!20}}

\ijcnlpfinalcopy

\title{Character-based Joint Segmentation and POS Tagging for Chinese \\ using Bidirectional RNN-CRF}

\author{Yan Shao \and Christian Hardmeier \and J{\"o}rg Tiedemann$^*$ \and Joakim Nivre \\
Department of Linguistics and Philology, Uppsala University \\
$^*$Department of Modern Languages, University of Helsinki \\
  {\tt \{yan.shao, christian.hardmeier, joakim.nivre\}@lingfil.uu.se} \\
  {\tt jorg.tiedemann@helsinki.fi}}

\date{}

\begin{document}

\maketitle

\begin{abstract}
We present a character-based model for joint segmentation and POS tagging for Chinese. The bidirectional RNN-CRF architecture for general sequence tagging is adapted and applied with novel vector representations of Chinese characters that capture rich contextual information and sub-character level features. The proposed model is extensively evaluated and compared with a state-of-the-art tagger respectively on CTB5, CTB9 and UD Chinese. The experimental results indicate that our model is accurate and robust across datasets in different sizes, genres and annotation schemes. We obtain state-of-the-art performance on CTB5, achieving 94.38 F1-score for joint segmentation and POS tagging. 
\end{abstract}

\section{Introduction}

Word segmentation and part-of-speech (POS) tagging are core steps for higher-level natural language processing (NLP) tasks. Given the raw text, segmentation is applied at the very first step and POS tagging is performed on top afterwards. As by convention the words in Chinese are not delimited by spaces, segmentation is non-trivial, but its accuracy has a significant impact on POS tagging. Moreover, POS tags provide useful information for word segmentation. Thus, modelling word segmentation and POS tagging jointly can outperform the pipeline models \cite{ng2004chinese, zhang2008joint}.  

POS tagging is a typical sequence tagging problem over segmented words, while segmentation also can be modelled as a character-level tagging problem via predicting the labels that identify the word boundaries. \newcite{ng2004chinese} propose a joint model which predicts the combinatory labels of segmentation boundaries and POS tags at the character level. Joint segmentation and POS tagging becomes a standard character-based sequence tagging problem and therefore the general machine learning algorithms for structured prediction can be applied.

The bidirectional recurrent neural network (RNN) using conditional random fields (CRF) \cite{lafferty2001conditional} as the output interface for sentence-level optimisation (BiRNN-CRF) achieves state-of-the-art accuracies on various sequence tagging tasks \cite{huang2015bidirectional, maend} and outperforms the traditional linear statistical models. RNNs with gated recurrent cells, such as long-short term memory (LSTM) \cite{hochreiter1997long} and gated recurrent units (GRU) \cite{cho2014properties} are capable of capturing long dependencies and retrieving rich global information. The sequential CRF on top of the recurrent layers ensures that the optimal sequence of tags over the entire sentence is obtained.  

In this paper, we model joint segmentation and POS tagging as a fully character-based sequence tagging problem via predicting the combinatory labels. The BiRNN-CRF architecture is adapted and applied. The Chinese characters are fed into the neural networks as vector representations. In addition to utilising the pre-trained character embeddings, we propose a concatenated n-gram-representation of the characters. Furthermore, sub-character level information, namely radicals and orthographical features extracted by convolutional neural networks (CNNs), are also incorporated and tested. Three datasets of different sizes, genres and with different annotation schemes are employed for evaluation. Our model is thoroughly evaluated and compared with the joint segmentation and POS tagging model in ZPar \cite{zhang2010fast}, which is a state-of-the-art joint tagger using structured perceptron and beam decoding. According to the experimental results, our proposed model outperforms ZPar on all the datasets in terms of accuracy. 

The main contributions of this work include: \begin{enumerate*} 
\item We apply the BiRNN-CRF model for general sequence tagging to joint segmentation and POS tagging for Chinese and achieve state-of-the-art accuracy. The experimental results show that our tagger is robust and accurate across datasets of different sizes, genres and annotation schemes.  
\item We propose a novel approach for vector representations of characters that leads to substantial improvements over the baseline model.
\item Additional improvements are obtained via exploring the feasibility of utilising sub-character level information. 
\item We provide an open-source implementation of our method along with pre-trained character embeddings.\footnote{
https://github.com/yanshao9798/tagger
}
\end{enumerate*}
\section{Model}

\subsection{Neural Network Architecture}

\begin{figure}
\scalebox{0.78}{
\begin{tikzpicture}[x=1.5cm, y=1.5cm]

\node (char1) {夏};
\node (char2) [right=1cm of char1]{天};
\node (char3) [right=1cm of char2]{太};
\node (char4) [right=1cm of char3]{热};

\node (eng2) [above=0.2cm of char3]{(too)};
\node (eng3) [above=0.2cm of char4]{(hot)};
\node (eng1) [left=1cm of eng2]{(summer)};

\foreach \x in {1,...,4}
	\node [circle, draw, minimum size=0.8cm] (emb1-\x) [below=0.2cm of char\x]{};
\foreach \x in {1,...,4}	
	\node [circle, draw, minimum size=0.8cm] (emb2-\x) [below=0.05cm of emb1-\x]{};
\foreach \x in {1,...,4}	
	\node [circle, draw, minimum size=0.8cm] (emb3-\x) [below=0.05cm of emb2-\x]{};
\foreach \x in {1,...,4}	
	\node[fit=(emb1-\x)(emb3-\x),inner sep=0, draw](emb-\x) {};

\node (note1) [left=0.3cm of emb2-1, text width=2cm, align=center]{character \\ representations};

\foreach \x in {1,...,4}
	\node [circle, draw, minimum size=0.8cm] (fgru-\x) [below=1cm of emb-\x]{GRU};	

\coordinate  [draw=none, left=0.6cm of fgru-1](fgru-0) ;
\coordinate  [draw=none, right=0.6cm of fgru-4](fgru-5);

\foreach \x [count=\xi from 1] in {0,...,4}
	\draw [-{Latex[length=2mm]}] (fgru-\x) -- (fgru-\xi);
 
\foreach \x in {1,...,4}
	\draw [-{Latex[length=2mm]}] (emb3-\x) -- (fgru-\x);

\node (note2) [left=0.3cm of fgru-1, text width=2cm, align=center]{forward\\ RNN};

\foreach \x in {1,...,4}
	\node [circle, draw, minimum size=0.8cm] (bgru-\x) [below=1cm of fgru-\x]{GRU};

\coordinate  [draw=none, left=0.6cm of bgru-1](bgru-0) ;
\coordinate  [draw=none, right=0.6cm of bgru-4](bgru-5);

\foreach \x [count=\xi from 1] in {0,...,4}
	\draw [{Latex[length=2mm]}-] (bgru-\x) -- (bgru-\xi);

\foreach \x in {1,...,4}
	\draw[-{Latex[length=2mm]}] (emb3-\x.south) to [out=210,in=135] (bgru-\x.north) ;

\node (note3) [left=0.3cm of bgru-1, text width=2cm, align=center]{backward\\ RNN};

\node [circle, draw, minimum size=0.8cm] (crf-1) [below=1cm of bgru-1]{B-NT};
\node [circle, draw, minimum size=0.8cm] (crf-2) [below=1cm of bgru-2]{E-NT};
\node [circle, draw, minimum size=0.8cm] (crf-3) [below=1cm of bgru-3]{S-AD};
\node [circle, draw, minimum size=0.8cm] (crf-4) [below=1cm of bgru-4]{S-VA};

\node (note4) [left=0.3cm of crf-1, text width=2cm, align=center]{CRF\\ Layer};

\foreach \x in {1,...,4}
	\draw [-{Latex[length=2mm]}] (bgru-\x) -- (crf-\x)[dashed];

\foreach \x in {1,...,4}
	\draw[-{Latex[length=2mm]}] (fgru-\x.south) to [out=335,in=40] (crf-\x.north)[dashed] ;

\foreach \x [count=\xi from 2] in {1,...,3}
	\draw [-] (crf-\x) -- (crf-\xi);

\node (post-3) [below=0.6cm of crf-3]{太\_AD};
\node (post-4) [below=0.6cm of crf-4]{热\_VA};
\node (post-12) [left=1cm of post-3]{夏天\_NT};

\node (note4) [left=0.95cm of post-12, text width=2cm, align=center]{Output};

\end{tikzpicture}
}
\caption{The BiRNN-CRF model for joint Chinese segmentation and POS tagging. The dashed arrows indicate that dropout layers are applied to the outputs of the recurrent layers.}\label{fig:1}
\end{figure}
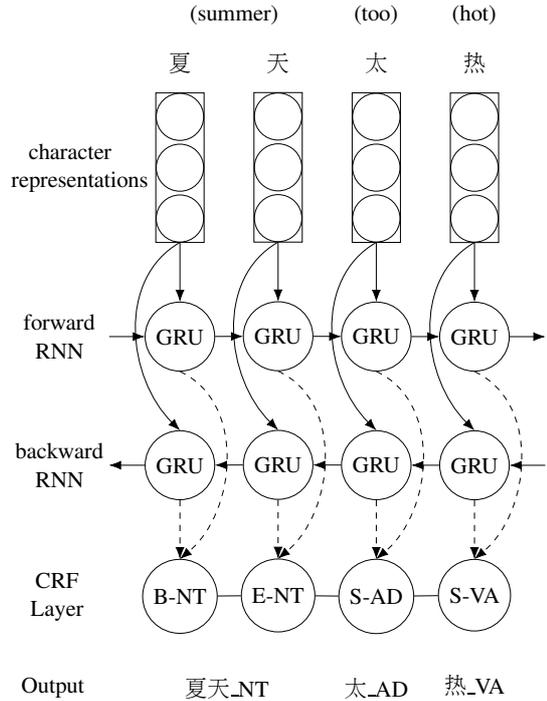

Our baseline model is an adaptation of BiRNN-CRF. As illustrated in Figure \ref{fig:1}, the Chinese characters are represented as vectors and fed into the bidirectional recurrent layers. The character representations will be described in detail in the following sections. For the recurrent layer, we employ GRU as the basic recurrent unit as it has similar functionalities but fewer parameters compared to LSTM \cite{chung2014empirical}. Dropout \cite{srivastava2014dropout} is applied to the outputs of the bidirectional recurrent layers. The outputs are concatenated and passed to the first-order chain CRF layer. The optimal sequence of the combinatory labels is predicted at the end. There is a post processing step to retrieve both segmentation and POS tags from the combinatory tags.

\begin{CJK}{UTF8}{gbsn}

\subsection{Tagging Scheme}

Following the work of \newcite{kruengkrai2009error}, the employed tags indicating the word boundaries are {\tt B}, {\tt I}, {\tt E}, {\tt S} representing a character at the beginning, inside, end of a word or as a single-character word. The CRF layer models conditional scores over all possible combinatory labels given the input characters. Incorporating the transition scores between the successive labels, the optimal sequence can be obtained efficiently via the Viterbi algorithm both for training and decoding. 

The time complexity for the Viterbi algorithm is linear with respect to the sentence length $n$ as $\mathcal{O}(k^2n)$, where $k$ is constant and equals to the total number of combinatory labels. The efficiency can be improved if we reduce $k$. For some POS tags, combining them with the full boundary tags is redundant. For instance, only the functional word  的 can be tagged as {\tt DEG} in Chinese Treebank \cite{xue2005penn}. Since it is a single-character word, combinatory tags of {\tt B-DEG}, {\tt I-DEG}, and {\tt E-DEG} never occur in the experimental data and should therefore be pruned to reduce the search space. Similarly, if the maximum length of words under a given POS tag is two in the training data, we prune the corresponding label.

\subsection{Character Representations}

We propose three different approaches to effectively represent Chinese characters as vectors for the neural network. 

\subsubsection{Concatenated N-gram}\label{sec1}

The prevalent character-based neural models assume that larger spans of text, such as words and n-grams, can be represented by the sequence of characters that they consist of. For example, the vector representation $V_{m, n}$ of a span $c_{m, n}$ is obtained by passing the vector representations $v_i$ of the characters $c_i$ to a functions $f$ as:
\begin{equation}\label{eq:1}
V_{m, n} = f(v_m, v_{m+1}, ..., v_n) 
\end{equation}
where $f$ is usually an RNN \cite{wangfind} or a CNN \cite{dos2014learning}.

\begin{figure}
\scalebox{0.78}{
\begin{tikzpicture}[x=1.5cm, y=1.5cm]

\node (char1) {夏};
\node (char2) [right=1.2cm of char1]{天};
\node (char3) [right=1.2cm of char2]{太};
\node (char4) [right=1.2cm of char3]{热};

\node (eng2) [above=0.4cm of char3]{(too)};
\node (eng3) [above=0.4cm of char4]{(hot)};
\node (eng1) [left=1.15cm of eng2]{(summer)};

\node[fit=(char3)(char3),inner sep=0, draw](char33)[dashed] {};
\node[fit=(char2)(char3),inner sep=0.1cm, draw](char23){};
\node[fit=(char2)(char4),inner sep=0.2cm, draw](char24)[dashed] {};

\node [circle, draw, minimum size=0.8cm] (gram1) [below=2cm of char2]{};
\node[fit=(gram1)(gram1),inner sep=0, draw](gram11){};

\node [circle, draw, minimum size=0.8cm] (gram2) [right=0.2cm of gram11]{};
\node [circle, draw, minimum size=0.8cm] (gram3) [right=0.05cm of gram2]{};
\node[fit=(gram2)(gram3),inner sep=0, draw](gram22){};

\node [circle, draw, minimum size=0.8cm] (gram4) [right=0.2cm of gram22]{};
\node [circle, draw, minimum size=0.8cm] (gram5) [right=0.05cm of gram4]{};
\node [circle, draw, minimum size=0.8cm] (gram6) [right=0.05cm of gram5]{};
\node[fit=(gram4)(gram6),inner sep=0, draw](gram33){};

\draw[-{Latex[length=2mm]}] (char33.south) to [out=210,in=95] (gram11.north);
\draw[-{Latex[length=2mm]}] (char23.south) to [out=190,in=95] (gram22.north);
\draw[-{Latex[length=2mm]}] (char24.south) to [out=270,in=95] (gram33.north);

\node (vec1) [below=0.4cm of gram11]{$V_{i, i}$};
\node (vec2) [below=0.4cm of gram22]{$V_{i-1, i}$};
\node (vec3) [below=0.4cm of gram33]{$V_{i-1, i+1}$};

\node (note1) [left=0.8cm of gram11, text width=2cm, align=center]{n-gram character\\ Representation $V_3$};

\end{tikzpicture}
}
\caption{Vector representations of the Chinese characters as incrementally concatenated n-gram vectors in a given context.}\label{fig:2}
\end{figure}
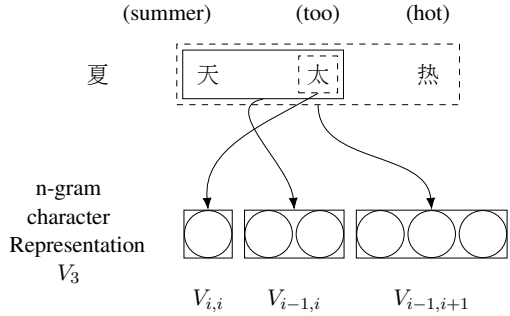

In this paper, instead of completely relying on the BiRNN to extract contextual features from context-free character representations, we encode rich local information in the character vectors via employing the incrementally concatenated n-gram representation as demonstrated in Figure \ref{fig:2}. In the example, the vector representation of the pivot character {\tt 太} in the given context is the concatenation of the context-free vector representation $V_{i, i}$ of {\tt 太} itself along with $V_{i-1, i}$ of the bigram {\tt 天太} as well as $V_{i-1, i+1}$ of the trigram {\tt天太热}.

Instead of constructing the vector representation $V_{m, n}$ of an n-gram $c_{m, n}$ from the character representations as in Equation \ref{eq:1}, $V_{m, n}$ in different orders,  such as $V_{i, i}$, $V_{i-1, i}$, and $V_{i-1, i+1}$,  are randomly initialised separately. We use a single special vector to represent all the unknown n-grams per order. The n-grams in different orders are then concatenated incrementally to form up the vector representations of a Chinese character in the given context, which is passed further to the recurrent layers. As shown in Figure \ref{fig:2}, the neighbouring characters on both sides of the pivot character are taken into account. 

\subsubsection{Radicals and Orthographical Features}

Chinese characters are logograms. As opposed to alphabetical languages, there is rich information encrypted in the graphical components. For instance, the Chinese characters that share the same part {\tt 钅} (gold) are all somewhat related to metals, such as {\tt 银} (silver), {\tt 铁} (iron), {\tt 针} (needle) and so on. The shared part {\tt 钅} is known as the radical, which functions as a semantic indicator. Hence, we investigate the effectiveness of using the information below the character level for our task. 

Radicals are first represented as randomly initialised vectors and concatenated as parts of the character representations. Radicals are traditionally used as indices in Chinese dictionaries. In our approach, they are retrieved via the unicode representation of Chinese characters as the characters that share the same radical are grouped together. They are organised in consistent with the categorisation in Kangxi Dictionary (康熙字典), in which all the Chinese characters are grouped under 214 different radicals. We only employ the radicals of the common characters in the unicode range of (U+4E00, U+9FFF). For the characters out of the range and the non-Chinese characters, we use a single special vector as their radical representations. 

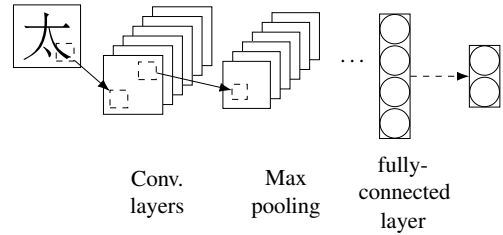
\begin{figure}
\scalebox{0.78}{
\begin{tikzpicture}[x=1.5cm, y=1.5cm]

\node [draw, minimum size=1.0cm](char1) {\Huge 太};

\node [draw, minimum size=0.3cm, below right=-0.6cm of char1](cell1)[dashed] {};

\node [draw, minimum size=1.0cm,  right=1.2cm of char1, fill=white](cnn1) {};

\foreach \x  [count=\xi from 2] in {1,...,5}
	\node [draw, minimum size=1.0cm, below left=-1.2cm of cnn\x, fill=white](cnn\xi) {};

\node[fit=(cnn1)(cnn6),inner sep=0](cnn){}; 

\node (note1) [below=0.8cm of cnn, text width=1.5cm, align=center]{Conv. layers};

\node [draw, minimum size=0.3cm, below left=-0.6cm of cnn6](cell2)[dashed] {};

\node [draw, minimum size=0.3cm, above right=-0.6cm of cnn6](cell3)[dashed] {};

\draw[-{Latex[length=2mm]}] (cell1) -- (cell2);

\foreach \x  in {1,...,6}
	\node [draw, minimum size=0.8cm, right=1 cm of cnn\x, fill=white](cnnp\x) {};

\node [draw, minimum size=0.15cm, below left=-0.6cm of cnnp6](cell4)[dashed] {};

\draw[-{Latex[length=2mm]}] (cell3) -- (cell4);

\node[fit=(cnnp1)(cnnp6),inner sep=0](out1){};

\node (note2) [right=0.4cm of note1, text width=1.5cm, align=center]{Max pooling};

\node [right= 0.2cm of out1](do1) {\ldots};

\node [circle, draw, minimum size=0.5cm] (o2) [right=1 cm of out1]{};

\node [circle, draw, minimum size=0.5 cm, above = 0.02cm of o2, fill=white](o1) {};

\foreach \x  [count=\xi from 3] in {2, 3}
	\node [circle, draw, minimum size=0.5 cm, below = 0.02cm of o\x, fill=white](o\xi) {};

\node[fit=(o1)(o4),inner sep=0, draw](out){};

\node (note3) [right=0.2cm of note2, text width=1.5cm, align=center]{fully-connected layer};

\node [circle, draw, minimum size=0.5cm] (fo1) [right=1 cm of o2]{};
\node [circle, draw, minimum size=0.5cm] (fo2) [right=1 cm of o3]{};

\node[fit=(fo1)(fo2),inner sep=0, draw](fout){};

\draw[-{Latex[length=2mm]}] (out) -- (fout)[dashed];

\end{tikzpicture}
}
\caption{Convolutional Neural Networks for orthographical feature extraction. Only the first convolutional layer and its following max-pooling layer are presented.}\label{fig:3}
\end{figure}

Additionally, instead of presuming that only radicals encode sub-character level information, we use convolutional neural networks (CNNs) to extract graphical features from scratch by regarding the Chinese characters as pictures and feed their pixels as the input. As illustrated in Figure \ref{fig:3},  there are two convolutional layers, both followed by a max-pooling layer. The output of the second max-pooling layer is reshaped and passed to a regular fully-connected layer. Dropout is applied to the output of the fully-connected layer. The output is then concatenated as parts of the character representation. The CNNs are trained jointly with the main network. 

\subsubsection{Pre-trained Character Embeddings}

The context-free vector representations of single characters introduced in section \ref{sec1} can be replaced by pre-trained character embeddings retrieved from large corpora. We employ GloVe \cite{pennington2014glove} to train our character embeddings on Wikipedia\footnote{https://dumps.wikimedia.org/} and the freely available Sogou News Corpora (SogouCS).\footnote{http://www.sogou.com/labs/resource/cs.php} We use randomly initialised vectors as the representations of the characters that are not in the embedding vocabulary. Pre-trained embeddings for higher-order n-grams are not employed in this paper.

\subsection{Ensemble Decoding}

At the final decoding phase, we use ensemble decoding, a simple averaging technique, to mitigate the deviations led by random weight initialisation of the neural network. For the chain CRF decoder, the final sequence of the combinatory tags $y$ is obtained via the conditional scores $S(y_i|x_i)$ and the transition scores $T(y_i, y_j)$ given the input sequence $x$. Instead of computing the optimal sequence with respect to the scores returned by a single model, both the conditional scores and transition scores are averaged over four models with identical parameter settings that are trained independently:
\begin{equation}
y^* = \argmax_{y \in L(x)} p(y|x; \bar{\{S\}}, \bar{\{T\}})
\end{equation}

Ensemble decoding is only applied to the best performing model according to the feature experiments at the final testing phase in this paper.

\section{Implementation}

Our neural networks are implemented using the TensorFlow 1.2.0 library \citep{abadi2016tensorflow}. We group the sentences with similar lengths into the same buckets and the sentences in the same bucket are padded to the same length accordingly. We construct sub-computational graphs respectively for each bucket. The training and tagging speed of our neural network on GPU devices can be drastically improved thanks to the bucket model. The training time is proportional to both the size of the training set and the number of POS tags.

\begin{table}
\centering
\scalebox{0.85}{
\begin{tabular}{l|c}
\hline
Char. embedding size & 64 \\
n-gram embedding size & 64 \\
Radical embedding size & 30 \\
\hline
Character font & simsun (宋体)\\
Character size & 30 $\times$ 30 \\
\hline
GRU state size & 200 \\
Conv. filter size & 5 $\times$ 5 \\
Conv. filter number & 32 \\
Max pooling size & 2 $\times$ 2 \\
Fully-connected size & 100 \\
\hline
Optimiser & Adagrad \\
Initial learning rate & 0.1 \\
Decay rate & 0.05 \\
Gradient Clipping & 5.0 \\
\hline
Dropout rate & 0.5 \\
Batch size & 10 \\
\hline
\end{tabular}
}
\caption{Hyper-parameters.}\label{tab:1}
\end{table}

Table \ref{tab:1} shows the adopted hyper-parameters. We use one set of parameters for all the experiments on different datasets. The weights of the neural networks, including the randomly intialised embeddings, are initialised using the scheme introduced in \newcite{glorot2010understanding}. The network is trained with the error back-propagation algorithm. All the embeddings are fine-tuned during training by back-propagating gradients. Adagrad  \cite{duchi2011adaptive} with mini-batches is employed for optimisation with the initial learning rate $ \eta_0 = 0.1$, which is updated with a decay rate $\rho = 0.05$ as $\eta_t = \frac{\eta_0}{\rho (t - 1) + 1} $,  where $t$ is the index of the current epoch. 

The model is optimised with respect to the performance on the development sets. F1-scores of both segmentation ($F1_{Seg}$) and joint POS tagging ($F1_{Seg\&Tag}$) are employed as $F1_{Seg}*F1_{Seg\&Tag}$ to measure the performance of the model after each epoch during training. In our experiments, the models are trained for 30 epochs. To ensure that the weights are well optimised, we only adopt the best epoch after the model is trained at least for 5 epochs. 

\section{Experiments}

\subsection{Datasets}

We employ three different datasets for our experiments, namely Chinese Treebank \cite{xue2005penn} 5.0 (CTB5) and 9.0 (CTB9) along with the Chinese section in Universal Dependencies (UD Chinese) \cite{nivre2016universal} of version 1.4. 

CTB5 is the most employed dataset for joint segmentation and POS tagging in previous research. It is composed of newswire data. We follow the conventional split of the dataset as in \newcite{jiang2008cascaded, kruengkrai2009error, zhang2010fast}. CTB9 consists of source texts in various genres, CTB5 is a subset of it. We split CTB9 by referring to the partition of CTB7 in \newcite{wang2011improving}. 
We extend the training, development and test sets from CTB5 by adding 80\% of the new data in CTB9 to training and 10\% each to development and test. The double-checked files are all placed in the test set. The detailed splitting information can be found in Table \ref{tab:a1} in Appendix. UD Chinese has both universal and language-specific POS tags. They are not predicted jointly in this paper. For the sake of convenience, we refer the universal tags as UD1 and the language-specific ones as UD2 in the following sessions. To make the model benefit from the pre-trained character embeddings, we convert the texts in UD Chinese from traditional Chinese into simplified Chinese. 

Table \ref{tab:2} shows brief statistics of the employed datasets in numbers of words. The out-of-vocabulary (OOV) words are counted regardless of the POS tags. We can see that the size of UD Chinese is much smaller and it has a notably higher OOV rate than the two CTB datasets. 

\begin{table}[h!]
\centering
\scalebox{0.88}{
\begin{tabular}{l|ccc}
\hline
 & CTB5 & CTB9 & UD Chinese \\
 \hline
 Train & 493,935 & 1,696,322 & 98,608 \\
 Dev & 6,821 & 136,468 & 12,663 \\
 Test & 8,008 & 242,317 & 12,012 \\
 \hline
OOV rate (dev) & 8.11 & 2.93 & 12.13 \\
OOV rate (test) & 3.47 & 3.13 & 12.46 \\
\hline
\end{tabular}
}
\caption{Statistics of the employed datasets in numbers of words.}\label{tab:2}
\end{table}

\begin{table*}[!htbp]
\centering
\scalebox{0.86}{
\begin{tabular}{c|cc|cc|cc|cc}
\hline
 & \multicolumn{2}{c|}{CTB5} & \multicolumn{2}{c|}{CTB9} & \multicolumn{2}{c|}{UD1} & \multicolumn{2}{c}{UD2} \\
\hline
& Seg & Seg\&Tag & Seg & Seg\&Tag & Seg& Seg\&Tag & Seg& Seg\&Tag  \\
\hline
size = 300 & 95.22 & 91.71 & 95.53 & 90.89 & 91.84 & 85.43 & 92.40 & 85.63  \\
\hline
1-gram & 95.14 & 91.52 & 95.25 & 90.43 & 91.74 & 85.07 & 91.83 & 84.93  \\
2-gram & 97.08 & 93.72 & 96.30 & 91.66 & \textbf{94.50} & 88.36 & 94.42 & 88.14 \\
3-gram & \textbf{97.14} & 94.01 & 96.47 & 91.75 & 94.36 & 88.27 & \textbf{94.43} & \textbf{88.32} \\
4-gram & 97.13 & \textbf{94.02} & 96.48 & \textbf{91.89} & 94.25 & 88.37 & 94.16 & 88.24 \\
5-gram & 96.94 & 93.84 & \textbf{96.50} & 91.88 & 94.40 & \textbf{88.47} & 94.25 & 88.03 \\
\hline
\end{tabular}
}
\caption{Evaluation of concatenated n-gram representations on the development sets in F1-scores}\label{tab:3}
\end{table*}

\begin{table*}[!htbp]
\centering
\scalebox{0.86}{
\begin{tabular}{c|cc|cc|cc|cc}
\hline
 & \multicolumn{2}{c|}{CTB5} & \multicolumn{2}{c|}{CTB9} & \multicolumn{2}{c|}{UD1} & \multicolumn{2}{c}{UD2} \\
\hline
& Seg& Seg\&Tag & Seg& Seg\&Tag & Seg& Seg\&Tag & Seg& Seg\&Tag  \\
\hline
3-gram & 97.14 & 94.01 & 96.47 & 91.75 & 94.36 & 88.27 & \textbf{94.43} & 88.32 \\
\hline
+radicals & \textbf{97.26} & \textbf{94.42} & 96.42  & 91.74  & 94.37 & 88.21 & 94.39 & \textbf{88.36}  \\
+graphical & 97.25 & 94.08 & \textbf{96.50}  & \textbf{91.78}  & \textbf{94.50} & \textbf{88.59} & 94.23  & 87.95  \\
\hline
\end{tabular}
}
\caption{Evaluation of sub-character level features on the development sets in F1-scores. }\label{tab:4}
\end{table*}

\begin{table*}[!htbp]
\centering
\scalebox{0.86}{
\begin{tabular}{c|cc|cc|cc|cc}
\hline
 & \multicolumn{2}{c|}{CTB5} & \multicolumn{2}{c|}{CTB9} & \multicolumn{2}{c|}{UD1} & \multicolumn{2}{c}{UD2} \\
\hline
& Seg& Seg\&Tag & Seg& Seg\&Tag & Seg& Seg\&Tag & Seg& Seg\&Tag  \\
\hline
1-gram & 95.14 & 91.52 & 95.25 & 90.43 & 91.74 & 85.07 & 91.83 & 84.93  \\
\hline
+GloVe & 95.82 & 92.45 & 95.44 & 90.57 & 92.77 & 86.48 & 93.01 & 86.48 \\
\hline
3-gram, radicals& 97.26 & 94.42 &   96.42  & 91.74  & 94.37 & 88.21 & 94.39 & 88.36 \\
\hline
+GloVe & \textbf{97.42}  & \textbf{94.58} & \textbf{96.56} & \textbf{91.96} & \textbf{95.12}  & \textbf{89.69} & \textbf{95.02} & \textbf{89.20}  \\
\hline
\end{tabular}
}
\caption{Evaluation of the pre-trained character embeddings on the development sets in F1-scores. }\label{tab:5}
\end{table*}

\subsection{Experimental Results}

Both segmentation (Seg) and joint segmentation and POS tagging (Seg\&Tag) are evaluated in our experiments.\footnote{The evaluation script is downloaded from: \\http://people.sutd.edu.sg/~yue\_zhang/doc/doc/joint\_files\\/evaluate.py} 
We employ word-level recall (R), precision (P) and F1-score (F) as the evaluation metrics. A series of feature experiments are carried out on the development sets to evaluate the effectiveness of the proposed approaches for vector representations of the characters. Finally, the best performing model according to the feature experiment is applied to the test sets in the forms of single as well as ensemble and compared with ZPar.  

\subsubsection{Feature Experiments}

Table \ref{tab:3} shows the evaluation results of using concatenated n-grams up to different orders as the character representations. By introducing 2-grams, we can obtain vast improvements over solely using the conventional character embeddings, which indicates that not all the local information can be effectively captured by the BiRNN using context-free character representations. Utilising the concatenated n-grams ensures that the same character has different but yet closely related representations in different contexts, which is an effective way to encode contextual features.

From the table, we see that notable improvements can be achieved further via employing 3-grams. 4-grams still help but only to CTB9 while adding 5-grams achieves almost no improvement on any of the datasets. The results imply that concatenating higher-order n-grams can be detrimental, especially on datasets in smaller sizes due to the fact that higher-order n-grams are more sparse in the training data and their vector representations cannot be trained well enough. Besides, adopting higher-order n-grams also substantially increases the numbers of weights and therefore both training and decoding become less efficient. Under the circumstances, we consider that 3-gram model is optimal for our task and it is employed in the following experiments for all the datasets.  

The concatenated n-grams have a bigger size compared to the basic character representation. We conduct one additional experiment using a basic 1-gram character model with a larger character vector size of 300. The evaluation scores are similar to the basic character model with the size of 64, which shows that the improvements obtained by the n-gram model are not matched by enlarging the size of the vector representation. 

The evaluation scores of the sub-character level features are reported in Table \ref{tab:4}. The relevant features are added on top of the 3-gram model. Employing radicals and graphical features achieves similar improvements for segmentation while utilising radicals obtains better results for joint POS tagging on CTB5. However, radicals are not a very effective feature on CTB9, UD1 and UD2 whereas a notable enhancement is observed when employing graphical features on UD1. Using CNNs to extract graphical features is computationally much more expensive than simply adopting radicals via a lookup table, especially when GPU is not available. 

From Table \ref{tab:5}, we can learn that employing pre-trained embeddings as initial vector representations for the characters achieves improvements in general, whereas the improvements are comparatively smaller if the the concatenated n-gram representations and the radicals are added. Additionally, the improvements obtained on UD Chinese are more significant than on CTBs, which indicates that the pre-trained character embeddings are more beneficial to the datasets in smaller sizes. 

In general, the feature experiments indicate that the proposed Chinese character representations are all sensitive to dataset size. Using higher-order n-grams requires more data for training. On the other hand, the pre-trained embeddings are more vital if the dataset is small. In addition, the different representations are sensitive to tagging schemes as the evaluation results on UD1 and UD2 are quite diverse. Taking both robustness and efficiency into consideration, we select 3-grams along with radicals and pre-trained character embeddings as the best setting for final evaluation. 

\subsubsection{Final Results}

Table \ref{tab:6} shows the final scores on the test sets. The complete evaluation results in precision, recall and F1-scores are contained in Table \ref{tab:a2} and Table \ref{tab:a3} in Appendix. Our system is compared with ZPar. We retrained a ZPar model on CTB5 that reproduces the evaluation scores reported in \newcite{zhang2010fast}. We also modified the source code so that it is applicable to CTB9 and UD Chinese. In addition, we perform the mid-$p$ McNemar's test \cite{fagerland2013mcnemar} to examine the statistical significances.

\begin{table*}[!htbp]
\centering
\scalebox{0.84}{
\begin{tabular}{c|c|ll|ll|ll|ll}
\hline
 \multicolumn{2}{c|}{} & \multicolumn{2}{c|}{CTB5} & \multicolumn{2}{c|}{CTB9} & \multicolumn{2}{c|}{UD1} & \multicolumn{2}{c}{UD2} \\
\hline
\multicolumn{2}{c|}{} & Seg& Seg\&Tag & Seg& Seg\&Tag & Seg& Seg\&Tag & Seg& Seg\&Tag  \\
\hline
\multicolumn{2}{c|}{ZPar} & 97.77 & 93.82 & 96.28 & 91.62 & 93.75 & 88.11 & 93.98 & 88.16  \\
\hline
\multicolumn{2}{c|}{Single (3-gram, rad., GloVe)} & 97.89  & 94.07**  & 96.47**  & 91.89**  & 94.85** & 89.41** & 94.93**  & 89.00**  \\
\multicolumn{2}{c|}{Ensemble (4 models)} & \textbf{98.02}* & \textbf{94.38}** & \textbf{96.67}**  & \textbf{92.34}**  & \textbf{95.16}**  & \textbf{89.75}* & \textbf{95.09}*  & \textbf{89.42}**  \\
\hline
\hline
\multirow{3}{*}{OOV recall} & ZPar & 76.98 & 68.34 & \textbf{75.83} & 63.71 & 78.69 & 64.40 & 79.56 & 64.86 \\
\cline{2-10}
& Single & \textbf{78.78}  & 69.78  & 74.16  & 62.58  & 81.36  & 67.40  & 81.16 &  66.73 \\
& Ensemble & 77.34 & \textbf{70.50}  & 75.52  & \textbf{64.14}  & \textbf{82.16}  & \textbf{68.14} & \textbf{81.56} & \textbf{68.00}  \\
\hline
\end{tabular}

}
\caption{Evaluations of the best model on the final test sets in F1-scores as well as the recalls of out-of-vocabulary words. 
Significance tests for Single are in comparison to ZPar, while tests for Ensemble are in comparison to Single (\textsuperscript{**}$p<0.01$, 
 \textsuperscript{*}$p<0.05$) 
 }\label{tab:6}
\end{table*}

As shown in Table \ref{tab:6}, the single model is worse than the ensemble model but still outperforms ZPar on all the tested datasets. ZPar incorporates discrete local features at both character and word levels and employs structured perceptron for global optimisation, whereas we encode rich local information in the character representations and employ BiRNN to effectively extract global features and capture long term dependencies. The chain CRF layer is used for sentence-level optimisation, which functions similarly to structured perceptron. As opposed to the taggers built with traditional machine learning algorithms, our model avoids heavy feature engineering and benefits from large plain texts via utilising pre-trained character embeddings. It is also very flexible to add sub-character level features as parts of the character representations. The model performs very well despite being fully character based. Moreover, it has clear advantages when applied to smaller datasets like UD Chinese, while the prevalence is much smaller on CTB5.

Both our model and ZPar segment OOV words in UD Chinese with higher accuracies than the ones in CTBs despite that UD Chinese is notably smaller and the overall OOV rate is higher. Compared to CTB, the words in UD Chinese are more fine-grained and the average word length is shorter, which makes it easier for the tagger to correctly segment the OOV words as \newcite{zhang2016trans} show that the longer words are more difficult to be segmented correctly. For joint POS tagging for OOV words, the two systems both perform significantly better on CTB5 as it is only composed of news text. 

In general, our model is more robust to OOV words than ZPar, except that ZPar yields better result for segmentation by a small margin on CTB9. ZPar also obtains higher accuracy for joint POS tagging than the single model on CTB9. The differences between ZPar and our model for both segmentation and POS tagging are more substantial on UD Chinese, which indicates that our model is relatively more advantageous for handling OOV words when the training sets are small, whereas ZPar is able to perform equally well when substantial amount of training data is available as they achieve similar results on the CTB sets. 

The single model is further improved by ensemble-averaging four independently trained models. The improvements are not drastic but they are observed systematically across all the datasets. In general, ensemble decoding is beneficial to handling OOV words as well except that a small drop for segmentation on CTB5 is observed. 

\begin{table}[h!]
\centering
\scalebox{0.87}{
\begin{tabular}{c|ll|cc}
\hline
 & \multicolumn{2}{c|}{Ensemble} &  \multicolumn{2}{c}{ZPar}  \\
\hline
 & Seg& Seg\&Tag & Seg& Seg\&Tag \\
 \hline
  BN & \textbf{97.89}* & \textbf{94.48}** & 97.68 & 94.22 \\
  CS & \textbf{96.67}** & \textbf{91.78}** & 95.61 & 90.15 \\
  FM & \textbf{96.54}** & \textbf{91.92}** & 96.30 & 91.51 \\
  MG & \textbf{94.54}** & \textbf{89.23}** & 94.22 & 88.60 \\
  NS & \textbf{97.56} & \textbf{93.92}** & 97.49 & 93.70 \\
  SM & \textbf{96.43}** & \textbf{91.78}** & 96.13 & 90.32 \\
  SP & \textbf{97.29}** & \textbf{93.93}** & 96.69 & 93.35 \\
  WB & \textbf{94.27}** & \textbf{88.44}** & 93.38 & 86.88 \\
\hline
\end{tabular}
}
\caption{Evaluation on Broadcast News (BN), Conversations (CS), Forum (FM), Magazine (MG), News (NS), Short Messages (SM), Speech (SP) and Weblogs (WB) in CTB9. (\textsuperscript{**}$p<0.01$,  \textsuperscript{*}$p<0.05$) 
}\label{tab:7}
\end{table}

Table \ref{tab:7} displays the evaluation of the ensemble model and ZPar on the decomposed test sets of CTB9 in different genres. Our model surpasses ZPar on all the genres in both segmentation and joint POS tagging. The differences are subtle on the genres in which the texts are normalised, such as News and Broadcast News. This, to a very large extent, explains why our model is only marginally better than ZPar on CTB5, whereas the experimental results reveal that our model is substantially better at processing non-standard text as it yields significantly higher scores on Conversations, Short Messages and Weblogs. The evaluation results of both our model and ZPar vary substantially across different genres as some genres are fundamentally more challenging to process. 


Our models are compared with the previous best-performing systems on CTB5 in Table \ref{tab:8}. Our models are not optimised particularly with respect to CTB5 but still yield competitive results, especially for joint POS tagging. We are the first to report evaluation scores on CTB9 and UD Chinese.

\begin{table}[!htbp]
\centering
\scalebox{0.98}{
\begin{tabular}{c|cc}
\hline
& Seg& Seg\&Tag  \\
\hline
\newcite{kruengkrai2009joint} & 97.98  & 94.00   \\
\newcite{zhang2010fast} & 97.78 & 93.67 \\
\newcite{sun2011stacked} & \textbf{98.17} & 94.02 \\
\newcite{wang2011improving} & 98.11 & 94.18 \\
\newcite{shen2014} & 98.02 & 93.80 \\
\hline
Single & 97.89 & 94.07 \\
Ensemble & 98.02 & \textbf{94.38} \\
\hline
\end{tabular}
}
\caption{Result comparisions on CTB5 in F1-scores. }\label{tab:8}
\end{table}

\subsection{Tagging Speed}

Our joint segmentation and POS tagger is very efficient with GPU devices and can be practically used for processing very large files.  The memory demand of decoding is drastically milder compared to training, a large batch size therefore can be employed. The tagger takes constant time to build the sub-computational graphs and load the weights. 

With bucket size of 10 and batch size of 500, Table \ref{tab:81} shows the tagging speed of the tagger using a single Tesla K80 GPU card and the pre-trained model on CTB5. The tagging speed of ZPar is also presented for comparison. GPU devices are not supported by ZPar and therefore the tagging speed is calculated using an Intel Core i7 CPU.

\begin{table}[!htbp]
\centering
\scalebox{0.88}{
\begin{tabular}{c|c|rr}
\hline
& Init. Time (s)& Sentence/s & Chars/s  \\
\hline
Single & 20 & 299.40 & 40,188.17  \\
Ensemble & 23 & 230.41 & 30,928.22 \\
\hline
ZPar & 4 & 134.59 & 18,090.09 \\
\hline
\end{tabular}
}
\caption{Tagging speed in numbers of sentences and characters per second}\label{tab:81}
\end{table}

\section{Related Work}

The fundamental BiRNN-CRF architecture is task-independent and has been applied to many sequence tagging problems on Chinese. \newcite{peng2016multi} adopt the model for Chinese segmentation and named entity recognition in the context of multi-task and multi-domain learning. \newcite{dong2016character} employ a character level BiLSTM-CRF model that utilises radical-level information for Chinese named entity recognition. \newcite{ma2016new} use a similar architecture but feed the Chinese characters pair-wise as edge embeddings instead. Their model is applied respectively to chunking, segmentation and POS tagging. 

\newcite{zhengdeep} model joint Chinese segmentation and POS tagging via predicting the combinatory segmentation and POS tags. They employ the adaptation of  the feed forward neural network introduced in \newcite{collobert2011natural} that only extracts local features in a context window. A perceptron-style training algorithm is employed for sentence level optimisation, which is the same as the training algorithm of the BiRNN-CRF model. Their proposed model is not evaluated on CTB5 and therefore difficult to be compared with our system. \newcite{kong2015segmental} apply segmental recurrent neural networks to joint segmentation and POS tagging but the evaluation results are substantially below the state-of-the-art on CTB5.

\newcite{bojanowski2016enriching} retrieve word embeddings via representing words as a bag of character n-grams for morphologically rich languages. A similar character n-gram model is proposed by \newcite{wieting2016charagram}. \newcite{sun2014radical} attempt to encode radical information into the conventional character embeddings. The radical-enhanced embeddings are employed and evaluated for Chinese segmentation. The results show that radical-enhanced embeddings outperform both skip-ngram and continues bag-of-word \cite{mikolov2013efficient} in word2vec.

\section{Conclusion}

We adapt and apply the BiRNN-CRF model for sequence tagging in NLP to joint Chinese segmentation and POS tagging via predicting the combinatory tags of word boundaries and POS tags. Concatenated n-grams as well as sub-character features are employed along with the conventional pre-trained character embeddings as the vector representations for Chinese characters. The feature experiments indicate that concatenated n-grams contribute substantially. However, both radicals and graphical features as sub-character level information are less effective. How to incorporate the sub-character level information more effectively will be further explored in the future. 

The proposed model is extensively evaluated on CTB5, CTB9 and UD Chinese. Despite the fact that different character representation approaches are sensitive to data size and tagging schemes, we use one set of hyper-parameters and universal feature settings so that the model is robust across datasets. The experimental results on the test sets show that our model outperforms ZPar which is built on structured perceptron on all the datasets. We obtain state-of-the-art performances on CTB5. The results on UD Chinese and CTB9 also reveal that our model has great advantages in processing non-standard text, such as weblogs, forum text and short messages. Moreover, the implemented tagger is very efficient with GPU devices and therefore can be applied to tagging very large files.

\section*{Acknowledgments}

 We acknowledge the computational resources provided by CSC in Helsinki and Sigma2 in Oslo through NeIC-NLPL (www.nlpl.eu). This work is supported by the Chinese Scholarship Council (CSC) (No. 201407930015). 

\bibliography{ijcnlp2017}
\bibliographystyle{ijcnlp2017}

\newpage 

\section*{Appendix}

\begin{table}[h!]
\centering
\begin{tabularx}{0.5\textwidth}{l|X}
\hline
Dataset & CTB chapter IDs \\
\hline
Train &  0044-0143, 0170-0270, 0400-0899, 1001-1017, 1019, 1021-1035, 1037-1043, 1045-1059, 1062-1071, 1073-1117, 1120-1131, 1133-1140, 1143-1147, 1149-1151, 2000-2915, 4051-4099, 4112-4180, 4198-4368, 5000-5446, 6000-6560, 7000-7013 \\
\hline
Dev & 0301-0326, 2916-3030, 4100-4106, 4181-4189, 4369-4390, 5447-5492, 6561-6630, 7013-7014 \\
\hline
Test & 0001-0043, 0144-0169, 0271-0301, 0900-0931, 1018, 1020, 1036, 1044, 1060, 1061, 1072, 1118, 1119, 1132, 1141, 1142, 1148, 3031-3145, 4107-4111, 4190-4197, 4391-4411, 5493-5558, 6631-6700, 7015-7017 \\
\hline
\end{tabularx}
\caption{The split of Chinese Treebank 9.0}\label{tab:a1}
\end{table}

\begin{table}[h!]
\centering
\scalebox{0.95}{
\begin{tabular}{l|c|ccc}
\hline
 & & P & R & F \\
 \hline
 \multirow{2}{*}{CTB5} & Single  & 97.49 & 98.30 & 97.89 \\
& Ensemble & 97.57 & 98.47 & 98.02 \\
\hline
 \multirow{2}{*}{CTB9} & Single & 96.38 & 96.55 & 96.47 \\
 & Ensemble & 96.61 & 96.74 & 96.67 \\
 \hline
 \multirow{2}{*}{UD1} & Single & 94.71 & 94.99 & 94.85 \\
 & Ensemble & 95.07 & 95.27 & 95.17 \\
 \hline
 \multirow{2}{*}{UD2} & Single & 94.98 & 94.93 & 94.93 \\
 & Ensemble & 95.00 & 95.22 & 95.11 \\
 \hline
\end{tabular}
}
\caption{Evaluation of segmentations in precision, recall and F1-scores}\label{tab:a2}
\end{table}

\begin{table}[h!]
\centering
\scalebox{0.95}{
\begin{tabular}{l|c|ccc}
\hline
 & & P & R & F \\
 \hline
\multirow{2}{*}{CTB5} & Single & 93.68  & 94.47 & 94.07 \\
& Ensemble & 93.95 & 94.81 & 94.38 \\
 \hline
 \multirow{2}{*}{CTB9} & Single & 91.81  & 91.97 & 91.89 \\
& Ensemble & 92.28 & 92.40 & 92.34 \\
 \hline
 \multirow{2}{*}{UD1} & Single & 89.28 & 89.54 & 89.41 \\
& Ensemble & 89.67 & 89.86  & 89.77 \\
 \hline
 \multirow{2}{*}{UD2} & Single & 88.95 & 89.04  & 89.00 \\
 & Ensemble & 89.33  & 89.54 & 89.43 \\
 \hline
\end{tabular}
}
\caption{Evaluation of joint segmentations and POS tagging in precision, recall and F1-scores}\label{tab:a3}
\end{table}

\end{CJK}
\end{document}